\documentclass{article}

\usepackage{microtype}
\usepackage{graphicx}
\usepackage{subfig}
\usepackage{mathtools}
\usepackage{siunitx}
\usepackage{float}
\usepackage{placeins}
\usepackage{booktabs} % for professional tables

\usepackage{hyperref}

\usepackage[accepted]{icml2020}
\def\eqref#1{equation~\ref{#1}}
\def\1{\bm{1}}

\DeclareMathAlphabet{\mathsfit}{\encodingdefault}{\sfdefault}{m}{sl}
\SetMathAlphabet{\mathsfit}{bold}{\encodingdefault}{\sfdefault}{bx}{n}

\DeclareMathOperator*{\argmax}{arg\,max}
\DeclareMathOperator*{\argmin}{arg\,min}

\newcommand{\xxcomment}[4]{\textcolor{#1}{[$^{\textsc{#2}}_{\textsc{#3}}$ #4]}}
\renewcommand{\xxcomment}[4]{}

\newcommand{\E}[1]{\mathbb{E}\left[\,#1\,\right]}
\newcommand{\Ep}[2]{\mathbb{E}_{#1}\left[\,#2\,\right]}
\newcommand{\Vp}[2]{\mathbb{V}\mathrm{ar}_{#1}\left[#2\right] }
\newcommand{\I}{\mathbb{I}}

\icmltitlerunning{
    Controlling Overestimation Bias with Truncated Mixture of Continuous Distributional Quantile Critics 
}

\usepackage{tikz}
\usepackage{multirow}
\usetikzlibrary{positioning,angles,quotes}

\usepackage{amsthm}

\usepackage{svg}

\usepackage{amssymb}
\usepackage{siunitx}

\begin{document}

\twocolumn[
\icmltitle{
    Controlling Overestimation Bias with Truncated Mixture of Continuous Distributional Quantile Critics 
}



\icmlsetsymbol{equal}{*}

\begin{icmlauthorlist}
\icmlauthor{Arsenii Kuznetsov}{saic}
\icmlauthor{Pavel Shvechikov}{saic,hse}
\icmlauthor{Alexander Grishin}{saic,hselab}
\icmlauthor{Dmitry Vetrov}{saic,hselab}
\end{icmlauthorlist}

% \icmlaffiliation{saic}{Samsung AI center, Samsung Research Russia, Moscow, Russia}
\icmlaffiliation{saic}{Samsung AI center, Moscow, Russia}
\icmlaffiliation{hselab}{Samsung-HSE Laboratory, National Research University Higher School of Economics, Moscow, Russia}
\icmlaffiliation{hse}{National Research University Higher School of Economics, Moscow, Russia}

\icmlcorrespondingauthor{Arsenii Kuznetsov}{brickerino@gmail.com}

\icmlkeywords{Machine Learning, ICML}

\vskip 0.3in
]



\printAffiliationsAndNotice{}  % leave blank if no need to mention equal contribution

\newcommand{\method}{TQC}

\begin{abstract}

The overestimation bias is one of the major impediments to accurate off-policy learning. 
% One of the major impediments to accurate off-policy learning is the overestimation bias. 
This paper investigates a novel way to alleviate the overestimation bias in a continuous control setting. 
Our method---Truncated Quantile Critics, TQC,---blends three ideas: distributional representation of a critic, truncation of critics prediction, and ensembling of multiple critics.
Distributional representation and truncation allow for arbitrary granular overestimation control, while ensembling provides additional score improvements. 
TQC outperforms the current state of the art on all environments from the continuous control benchmark suite, demonstrating 25\% improvement on the most challenging Humanoid environment.
\end{abstract}

\section{Introduction}

Sample efficient off-policy reinforcement learning demands accurate approximation of the Q-function.
Quality of approximation is key for stability and performance, since it is the cornerstone for temporal difference target computation, and action selection in value-based methods \cite{mnih2013playing}, or policy optimization in continuous actor-critic settings \cite{haarnoja2018soft, fujimoto18a}. 

In continuous domains, policy optimization relies on gradients of the Q-function approximation, sensing and exploiting unavoidable erroneous positive biases. 
Recently, \citet{fujimoto18a} significantly improved the performance of a continuous policy by introducing a novel way to alleviate the overestimation bias \cite{thrun1993issues}.
We continue this line of research and propose an alternative highly competitive method for controlling overestimation bias.

\citet{thrun1993issues} elucidate the overestimation as a consequence of Jensen's inequality: the maximum of the Q-function over actions is not greater than the expected maximum of noisy (approximate) Q-function. 
    Specifically, for any action-dependent random noise $U(a)$ such that $ \forall a ~~ \Ep{U}{U(a)} = 0$,
    \begin{equation}
        \begin{aligned}
            \max_a Q(s,a) 
            &=   
            \max_a \Ep{U}{Q(s, a) + U(a)} \\  
            &\leq  
            \Ep{U}{\max_a \{ Q(s,a) + U(a) \} }.
        \end{aligned}
    \end{equation}
In practice, the noise $U(a)$ may arise for various reasons and from various sources, such as spontaneous errors in function approximation, Q-function invalidation due to ongoing policy optimization, stochasticity of environment, etc. 
Off-policy algorithms grounded in temporal difference learning are especially sensitive to approximation errors since errors are propagated backward through episodic time and accumulate over the learning process.

The de facto standard for alleviating overestimations in discrete control is the double estimator \cite{hasselt2010double,van2013estimating}. 
However, \citet{fujimoto18a} argue that for continuous control this estimator may still overestimate in highly variable state-action space regions, and propose to promote underestimation by taking the minimum over two separate approximators. 
These approximators constitute naturally an ensemble, the size of which controls the intensity of underestimation: more approximators correspond to more severe underestimation \cite{lan2020maxmin}.
We argue, that this approach, while very successful in practice, has a few shortcomings:
\begin{itemize}
    \item The overestimation control is coarse: it is impossible to take the minimum over a fractional number of approximators (see Section \ref{sec:single_state}).  
    \item The aggregation with $\min$ is wasteful: it ignores all estimates except the minimal one, diminishing the power of the ensemble of approximators.   
\end{itemize}
We address these shortcomings with a novel method called Truncated Quantile Critics (TQC).
In the design of TQC, we draw on three ideas:  distributional representation of a critic, truncation of approximated distribution, and ensembling.

\textbf{Distributional representations}~
The distributional perspective \cite{bellemare2017distributional} advocates the modeling of the \textit{distribution} of the random return, instead of the more common modeling of the Q-function, the expectation of the return.
In our work, we adapt QR-DQN \cite{dabney2018distributional} for continuous control and approximate the quantiles of the return distribution conditioned on the state and action.
Distributional perspective allows for learning the intrinsic randomness of the environment and policy, also called aleatoric uncertainty. 
We are not aware of any prior work employing aleatoric uncertainty for overestimation bias control.
We argue that the granularity of distributional representation is especially useful for precise overestimation control.

\begin{figure}[t]

\includegraphics[width=\columnwidth]{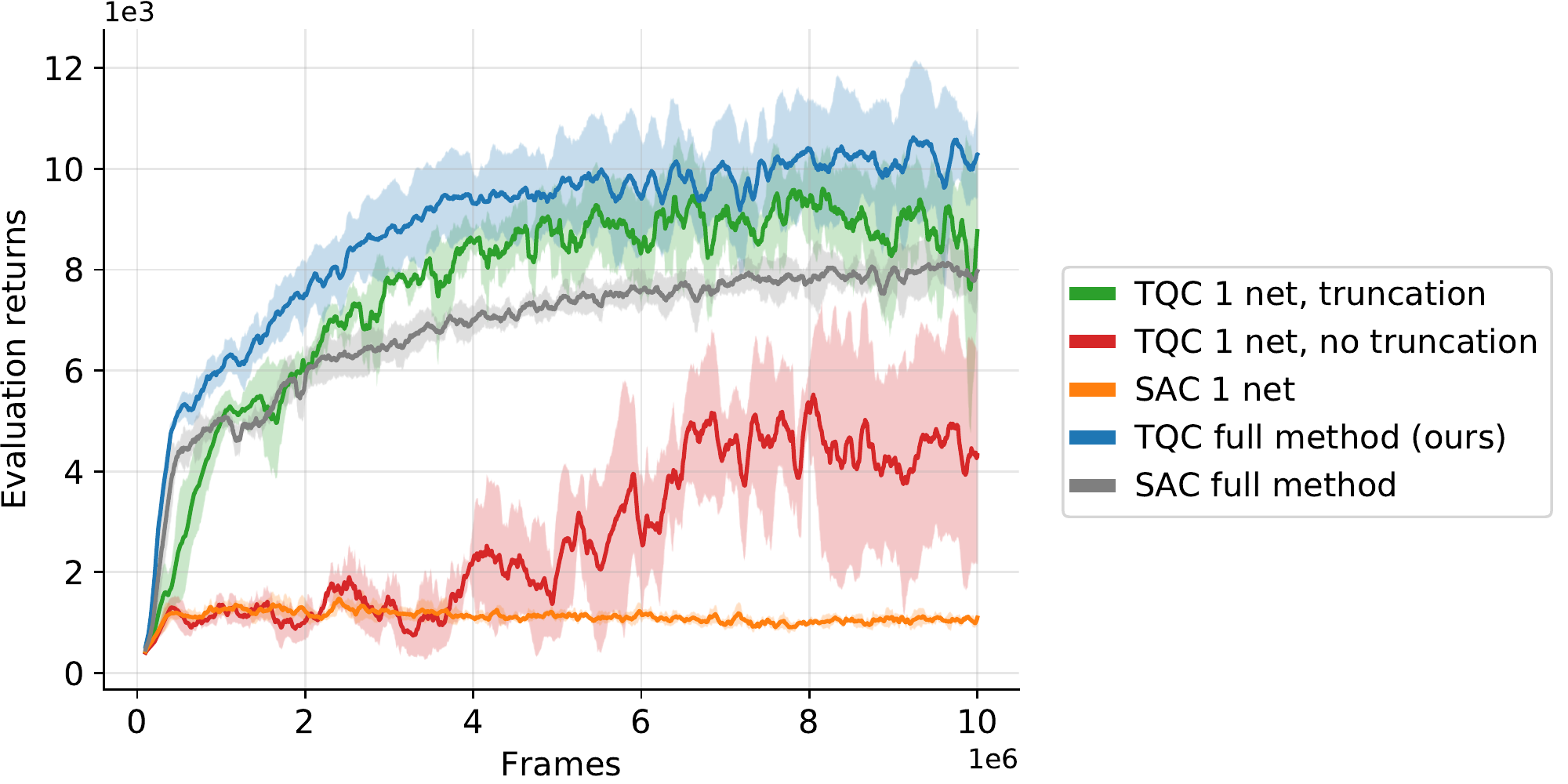}
\caption{
Evaluation on the Humanoid environment. Results are averaged over 4 seeds, $\pm$ std is shaded.
}
\label{fig:one_q_teaser}
% \vspace{-.5cm}
\end{figure}

\textbf{Truncation}~
To control the overestimation, we propose to truncate the right tail of the return distribution approximation by dropping several of the topmost atoms. 
By varying the number of dropped atoms, we can balance between over- and underestimation. 
In a sense, the truncation operator is parsimonious: we drop only a small number of atoms (typically, around 8\% of the total number of atoms).
Additionally, truncation does not require multiple separate approximators:  our method surpasses the current state of the art (which uses multiple approximators) on some benchmarks even using only a single one (Figure \ref{fig:one_q_teaser}).

\textbf{Ensembling}~
        The core operation of our method---truncation of return distribution---does not impose any restrictions on the number of required approximators.  
        This effectively decouples overestimation control from ensembling, which, in turn, provides for additional performance improvement (Figure~\ref{fig:one_q_teaser}).

Our method improves the performance on all environments in the standard OpenAI gym \cite{brockman2016openai} benchmark suite powered by MuJoCo \cite{todorov2012mujoco}, with up to 30\% improvement on some of the environments. 
For the most challenging Humanoid environment this improvement translates into twice the running speed of the previous SOTA (since agent gets $5$ as part of reward per step until it fell).  
The price to pay for this improvement is the computational overhead carried by distributional representations and ensembling (Section \ref{sec:hyperparms_sens}).

This work makes the following contributions to the field of continuous control:  
\begin{enumerate} 
    \item We design a practical method for the fine-grained control over the overestimation bias, called Truncated Quantile Critics (Section~\ref{sec:method}).
    For the first time, we (1) incorporate aleatoric uncertainty into the overestimation bias control,
    (2) decouple overestimation control and multiplicity of approximators, 
    (3) ensemble distributional approximators in a novel way.
    
    \item We advance the state of the art on the standard continuous control benchmark suite (Section~\ref{sec:experiments}) and perform extensive ablation study (Section~\ref{sec:ablation}).
\end{enumerate}

To facilitate reproducibility, we carefully document the experimental setup, perform exhaustive ablation, average experimental results over a large number of seeds,  publish raw data of seed runs, and release the code for Tensorflow\footnote{\url{https://github.com/bayesgroup/tqc}} and PyTorch\footnote{\url{https://github.com/bayesgroup/tqc_pytorch}}.

\section{Background}

\subsection{Notation}

We consider a  Markov decision process, MDP, defined by the tuple $(\mathcal{S}, \mathcal{A}, \mathcal{P}, R, \gamma)$, with continuous state and action spaces $\mathcal{S}$ and $\mathcal{A}$, unknown state transition density 
$\mathcal{P}: \mathcal{S} \times \mathcal{A} \times \mathcal{S} \rightarrow [0, \infty)$, 
random variable reward function $R$, 
and discount factor $\gamma \in [0, 1)$.

A policy $\pi$ maps each state $s \in \mathcal{S}$ to a distribution over $\mathcal{A}$. 
We write $\mathcal{H}(\pi(s_t))$ to denote the entropy of the policy conditioned on the state $s_t$.

We write  $\dim \mathcal{X} $ for the dimensionality of the space $\mathcal{X}$.
Unless explicitly stated otherwise, the $\Ep{\mathcal{D}, \pi}{\cdot}$ signifies the expectation over the $(s_t, a_t, r_t, s_{t+1})$ from experience replay $\mathcal{D}$, and $a_{t+1}$ from $\pi(\cdot | s_{t+1})$.
We use the overlined notation to denote the parameters of target networks, i.e., $\overline{\psi}$ denotes the exponential moving average of parameters~$\psi$.

\subsection{Soft Actor Critic}
\label{sec:sac}
The Soft Actor Critic (SAC) \cite{haarnoja2018soft} is an off-policy actor-critic algorithm based on the maximum entropy framework.
The objective  encourages policy stochasticity by augmenting the reward with the entropy at each step.

The policy parameters $\phi$ can be learned by minimizing the 
\begin{equation}\label{eq:sac_actor}
J_{\pi}(\phi)=
\Ep{\mathcal{D}, \pi}{
\mathrm{D}_{\mathrm{KL}}
    \left(
        \pi_{\phi} \left(\cdot | s_{t}\right) 
        \Big\| 
        \frac
        {\exp \left(\frac{1}{\alpha} Q_{\psi}\left(s_{t}, \cdot\right)\right)}
        {C_{\psi}\left(s_{t}\right)}
    \right)
},
\end{equation}
where $Q_\psi$ is the soft Q-function and $C_{\psi}\left(s_{t}\right) $ is the normalizing constant.

The soft Q-function parameters $\theta$ can be learned by minimizing the soft Bellman residual
\begin{equation}\label{eq:sac_critic}
    J_{Q}(\psi)=\mathbb{E}_{\mathcal{D}, \pi}
    \left[
        \frac{1}{2}
        \left(
            Q_{\psi}(s_{t}, a_{t})- y(s_t, a_t)
        \right)^{2}
    \right],
\end{equation}
where $y(s_t, a_t)$ denotes the temporal difference target
\begin{equation}
\label{eq:sac_target}
r(s_t, a_t) + \gamma \left[Q_{\overline \psi}(s_{t+1}, a_{t+1}) - \alpha \log \pi_\phi (a_{t+1}| s_{t+1})\right],
\end{equation}
and $\alpha$ is the entropy temperature coefficient. 
\citet{haarnoja2018applications} proposed to dynamically adjust the~$\alpha$ by taking a gradient step with respect to the loss 
\begin{equation}\label{eq:entropy_temp}
J(\alpha)= \mathbb{E}_{\mathcal{D}, \pi_\phi}
\left[
    \log \alpha \cdot  ( 
        -\log \pi_{\phi} ( a_{t} | s_{t} ) 
        -
        \mathcal{H}_T 
    ) 
\right], 
\end{equation}
each time the $\pi_\phi$ changes. 
This decreases the $\alpha$, if the stochastic estimate of policy entropy, $-\log \pi_\phi (a_t | s_t) $,  is higher than $\mathcal{H}_T$, and increases $\alpha$ otherwise.
The target entropy usually is set heuristically to  $\mathcal{H}_T = - \dim \mathcal{A}$.  

\citet{haarnoja2018applications} takes the minimum over two Q-function approximators to compute the target in \eqref{eq:sac_target} and policy objective in \eqref{eq:sac_actor}.

\subsection{Distributional Reinforcement Learning with Quantile Regression}
Distributional reinforcement learning focuses on approximating the return random variable 
$ Z^{\pi}(s, a) := \sum_{t=0}^{\infty} \gamma^{t} R\left(s_{t}, a_{t}\right)$ where 
$s_{0}=s$, $a_{0}=a$ and  $s_{t+1} \sim \mathcal{P}(\cdot | s_{t}, a_{t})$, $a_t \sim \pi(\cdot | s_t)$,
as opposed to approximating the expectation of the return, also known as the Q-function, 
    $Q^{\pi}(s, a):=  \E{Z^{\pi}(s, a)}$.

QR-DQN \cite{dabney2018distributional} approximates the distribution $Z^\pi(s,a)$ with 
$ Z_{\psi}(s, a):=\frac{1}{M} \sum_{m=1}^{M} \delta ( \theta^{m}_\psi(s, a) )$,
a mixture of atoms---Dirac delta functions at locations $\theta^1_\psi(s,a), \dots ,\theta^M_\psi(s,a)$ given by a parametric model  $\theta_{\psi}: \mathcal{S} \times \mathcal{A} \rightarrow \mathbb{R}^M$. 

Parameters $\psi$ are optimized by minimizing the averaged over the replay  1-Wasserstein distance between $Z_\psi$ and the temporal difference target distribution $\mathcal{T}_\pi Z_{\overline{\psi}}$,  where  $\mathcal{T}_\pi$ is the distributional Bellman operator \cite{bellemare2017distributional}:
\begin{equation}
    \begin{gathered}
        \mathcal{T}_\pi Z(s, a) :\stackrel{D}{=} R(s, a)+\gamma Z\left(s^{\prime}, a^{\prime}\right), \\
        s' \sim \mathcal{P}\left(\cdot | s, a\right), a' \sim \pi(\cdot | s').
    \end{gathered}
\end{equation}
As \citet{dabney2018distributional} show, this minimization can be performed by learning quantile locations for fractions $\tau_m = \frac{2m-1}{2M}, m \in [1..M]$ via  quantile regression.
The quantile regression loss, defined for a quantile fraction $\tau \in [0,1]$, is  
\begin{equation}
\begin{aligned} 
\mathcal{L}_{\mathrm{QR}}^{\tau}(\theta): &=\mathbb{E}_{\tilde{Z} \sim Z}\left[\rho_{\tau}(\tilde{Z}-\theta)\right], \text { where } \\ 
\rho_{\tau}(u) &=u\left(\tau-\I(u<0)\right), \forall u \in \mathbb{R}.
\end{aligned}
\end{equation}
To improve gradients for small $u$ authors propose to use the Huber quantile loss (asymmetric Huber loss): 
\begin{equation}  \label{eq:quant_huber}
\rho_{\tau}^{H}(u) = |\tau-\I(u<0)|  \mathcal{L}^{1}_H(u),
\end{equation}
where  $\mathcal{L}^{1}_H(u) $ is a Huber loss with parameter $1$.

\section{Truncated Quantile Critics, TQC}
\label{sec:method}

We start with an informal explanation of TQC and motivate our design choices.
Next, we outline the formal procedure at the core of TQC, specify the loss functions and present an algorithm for practical  implementation.  

\subsection{Overview}

\begin{figure*}[t]
\includegraphics[width=\textwidth]{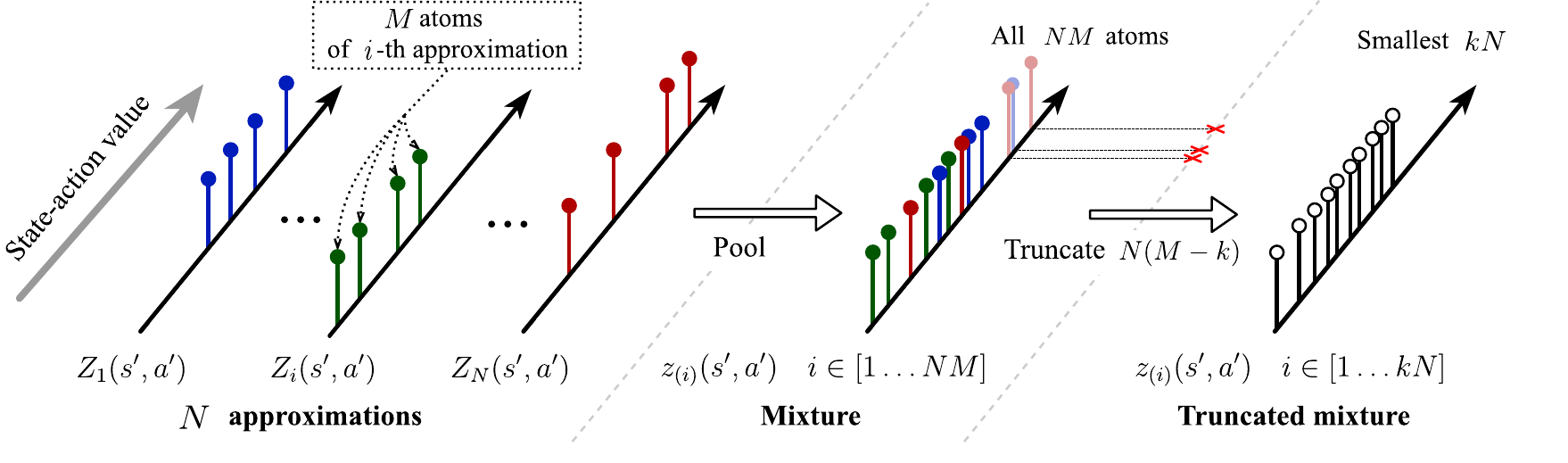}
\caption{
Selection of atoms for the temporal difference target distribution $Y(s,a)$. 
First, we compute approximations of the return distribution conditioned on $s'$ and $a'$ by evaluating  $N$ separate target critics. 
Second, we make a mixture out of the $N$ distributions from the previous step.
Third, we truncate the right tail of this mixture to obtain atoms $z_{(i)}(s',a')$ from \eqref{eq:tqc_atoms}.
}
\label{fig:scheme}
\end{figure*}

To achieve granularity in controlling the overestimation, we "decompose" the expected return into atoms of distributional representation. 
By varying the number of atoms, we can control the precision of the return distribution approximation. 

To control the overestimation, we propose to truncate the approximation of the return distribution:  we drop atoms with the largest locations and estimate the Q-value by averaging the locations of the remaining atoms. 
By varying the total number of atoms and the number of dropped ones, we can flexibly balance between under- and overestimation. 
The truncation naturally accounts for the inflated overestimation due to the high return variance:  the higher the variance, the lower the Q-value estimate after truncation. 

To improve the Q-value estimation, we ensemble multiple distributional approximators in the following way.
First, we form a mixture of distributions predicted by $N$ approximators. 
Second, we truncate this mixture by removing atoms with the largest locations and estimate the Q-value by averaging the locations of the remaining atoms. 
The order of operations---the truncation of the mixture vs. the mixture of truncated distributions---may matter. 
The truncation of a mixture removes the largest outliers from the pool of all predictions. 
Such a truncation may be useful in a hypothetical case of one of the critics goes crazy and overestimates much more than the others.
In this case, the truncation of a mixture removes the atoms predicted by this inadequate critic. 
In contrast, the mixture of truncated distributions truncates all critics evenly.

Our method is different from previous approaches \citep{zhang2019quota,dabney2018implicit} that distorted the critic's distribution at the policy optimization stage only.
We use nontruncated critics' predictions for policy optimization. And truncate target return distribution at the value learning stage. 
Intuitively, this prevents errors from propagating to other states via TD learning updates and eases policy optimization.  

Next, we present TQC formally and summarize the procedure in Algorithm \ref{alg:main}.

\subsection{Computation of the target distribution}

We propose to train $N$ approximations $Z_{\psi_1}, \dots Z_{\psi_N}$ of the policy conditioned return distribution $Z^\pi$.  
Each $Z_{\psi_n}$ maps each $(s, a)$ to a probability distribution 
\begin{equation}
    Z_{\psi_n}(s, a) := \frac{1}{M} \sum_{m=1}^{M} \delta \left(\theta_{\psi_n}^m(s, a) \right) ,
\end{equation}
supported on atoms $ \theta_{\psi_n}^1(s,a), \dots, \theta_{\psi_n}^M(s,a) $ .

We train approximations $Z_{\psi_1}, \dots Z_{\psi_N}$ on the temporal difference target distribution $Y(s,a)$.
We construct it as follows. 
We pool atoms of distributions $Z_{\psi_1}(s',a'), \dots, Z_{\psi_N}(s',a')$ into a set  
\begin{equation}
\mathcal{Z}(s',a') := \{\theta_{\psi_n}^m(s', a') \mid  n \in [1..N], m \in [1..M] \}
\end{equation}
and denote elements of $\mathcal{Z}(s',a')$ sorted in ascending order by $z_{(i)}(s',a')$, with $i \in [1..MN]$.

The $kN$ smallest elements of $\mathcal{Z}(s',a')$ define atoms 
\begin{equation}
\label{eq:tqc_atoms}
  y_i (s, a) := r(s,a) + \gamma [z_{(i)}(s', a')  - \alpha \log \pi_\phi(a' | s')]
\end{equation}
of the target distribution
\begin{equation}
Y(s,a) := \frac{1}{kN} \sum_{i=1}^{kN} \delta \left( y_i (s, a) \right) .
\end{equation}
In practice, we always populate $\mathcal{Z}(s',a')$ with atoms predicted by target networks $Z_{\overline{\psi}_1}(s',a'), \dots, Z_{\overline{\psi}_N}(s',a')$, which are more stable.

\subsection{Loss functions}

We minimize the 1-Wasserstein distance between each of $Z_{\psi_n}(s,a), n\in[1..N]$ and the temporal difference target distribution $Y(s,a)$. 
Equivalently \cite{dabney2018distributional}, to minimize this distance we can approximate the quantiles of the target distribution, i.e., learn the locations for quantile fractions $\tau_m = \frac{2m-1}{2M}, m \in [1..M]$.

We approximate the $\tau_m, m \in [1..M]$ quantiles of $Y(s,a)$ with  $ \theta_{\psi_n}^1(s,a), \dots, \theta_{\psi_n}^M(s,a) $  by minimizing the loss
\begin{equation}\label{eq:value_loss}
    J_Z (\psi_n) = 
    \Ep{\mathcal{D}, \pi} { 
             \mathcal{L}^k(s_t,a_t; \psi_n)
    }, 
\end{equation}
over the parameters $\psi_n$, where 
\begin{equation}
    \mathcal{L}^k(s,a; \psi_n) = \frac{1}{kNM} \sum_{m=1}^M \sum_{i=1}^{kN} \rho_{ \tau_m}^H(y_i(s, a) - \theta_{\psi_n}^m(s,a)).
\end{equation}
In this way, each learnable location $\theta_{\psi_n}^m(s,a)$ becomes dependent on all atoms of the truncated mixture of target distributions.

The policy parameters $\phi$ can be optimized to maximize the entropy penalized estimate of the Q-value by minimizing the loss 
\begin{equation}\label{eq:policy_loss}
    J_\pi(\phi) = 
    \Ep{\mathcal{D}, \pi} {
        \alpha \log \pi_\phi(a | s) 
        -
       \frac{1}{NM} 
       \sum_{m,n=1}^{M,N}
        \theta_{\psi_n}^m(s,a)
    }, 
\end{equation}
where $s \sim \mathcal{D}, a \sim \pi_\phi(\cdot | s)$. 
We use nontruncated estimate of the Q-value for policy optimization to avoid double truncation: Z-functions approximate already truncated future distribution.

\begin{algorithm}[tb]
   \caption{TQC.  $\hat \nabla$ denotes the stochastic gradient}
   \label{alg:main}
\begin{algorithmic}
    \STATE $\bullet$ Initialize policy $\pi_\phi$, critics $Z_{\psi_{n}}, Z_{\overline{\psi}_n}$ for $ n \in [1..N]$
    \STATE $\bullet$ Set  replay $\mathcal{D} = \varnothing$, $\mathcal{H}_T = - \dim \mathcal{A}$,  $\alpha =1$, $\beta=.005$
    \FOR{each iteration}
        \FOR{each environment step, until done}
            \STATE collect transition $(s_t, a_t, r_t, s_{t+1})$  with policy $\pi_\phi$
            \STATE $\mathcal{D} \leftarrow \mathcal{D} \cup \{ (s_t, a_t, r_t, s_{t+1}) \}$
        \ENDFOR
        \FOR{each gradient step}
            \STATE sample a batch from the replay $\mathcal{D}$
            \STATE $\alpha \leftarrow \alpha - \lambda_\alpha \hat \nabla_\alpha J(\alpha)$ 
            \hfill Eq. (\ref{eq:entropy_temp})
            \STATE $\phi \leftarrow \phi -  \lambda_\pi \hat \nabla_\phi J_\pi(\phi)$  
            \hfill Eq. (\ref{eq:policy_loss})
            \STATE  $\psi_n \leftarrow  \psi_n - \lambda_Z \hat \nabla_{\psi_n}  J_Z(\psi_n)$,  $n\in[1..N]$  
            \hfill Eq. (\ref{eq:value_loss})
            \vspace{1pt}
            \STATE  $\overline{\psi}_n \leftarrow  \beta \psi_n  + (1 - \beta) \overline{\psi}_n $, $n\in[1..N]$
        \ENDFOR
    \ENDFOR
    \STATE \textbf{return} policy $\pi_\phi$, critics $Z_{\psi_n}$, $n \in [1..N]$.
\end{algorithmic}
\end{algorithm}

\section{Experiments}
\label{sec:experiments}

First, we compare our method with other possible ways to mitigate the overestimation bias on a simple MDP, for which we can compute the true Q-function and the optimal policy.

Next, we quantitatively compare our method with competitors on a standard continuous control benchmark -- the set of MuJoCo \citep{todorov2012mujoco} environments implemented in OpenAI Gym \citep{brockman2016openai}.
The details of the experimental setup are in Appendix \ref{ap:exp_setting}.

We implement TQC on top of the SAC \citep{haarnoja2018applications} with auto-tuning of the entropy temperature (Section \ref{sec:sac}).
For all MuJoCo experiments, we use $N=5$ critic networks with three hidden layers of $512$ neurons each, $M=25$ atoms, and the best number of dropped atoms per network $d \in [0..5]$, if not stated otherwise. 
The other hyperparameters are the same as in SAC (see Appendix \ref{ap:hyperparams}).

\subsection{Single state MDP}
\label{sec:single_state}
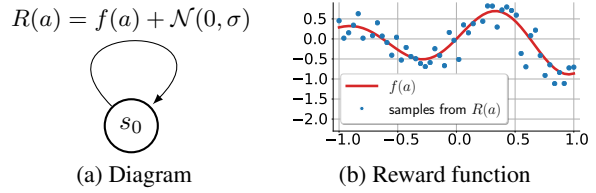
\begin{figure}[t]
\subfloat[Diagram]{  \begin{tikzpicture}[auto,node distance=8mm,>=latex]
    \tikzstyle{round}=[thick,draw=black,circle]
    \node[round] (s0) {$s_0$};
    % \node[round,above right=0mm and 20mm of s0] (s1) {$s_1$};
    % \node[round,below right=0mm and 20mm of s0] (s2) {$s_2$};
    % \draw[->] (s0) -- (s1);
    % \draw[->] (s0) -- (s2);
    \path[->] (s0) edge [out=135, in=45, loop] node {\footnotesize $R(a)=f(a) + \mathcal{N}(0, \sigma)$} (s0);
    % \path pic[draw, angle radius=6mm,angle eccentricity=1.2] {angle = s2--s0--aa};
\end{tikzpicture} } 
\subfloat[Reward function]{ 
        \includegraphics[width=0.5\columnwidth]{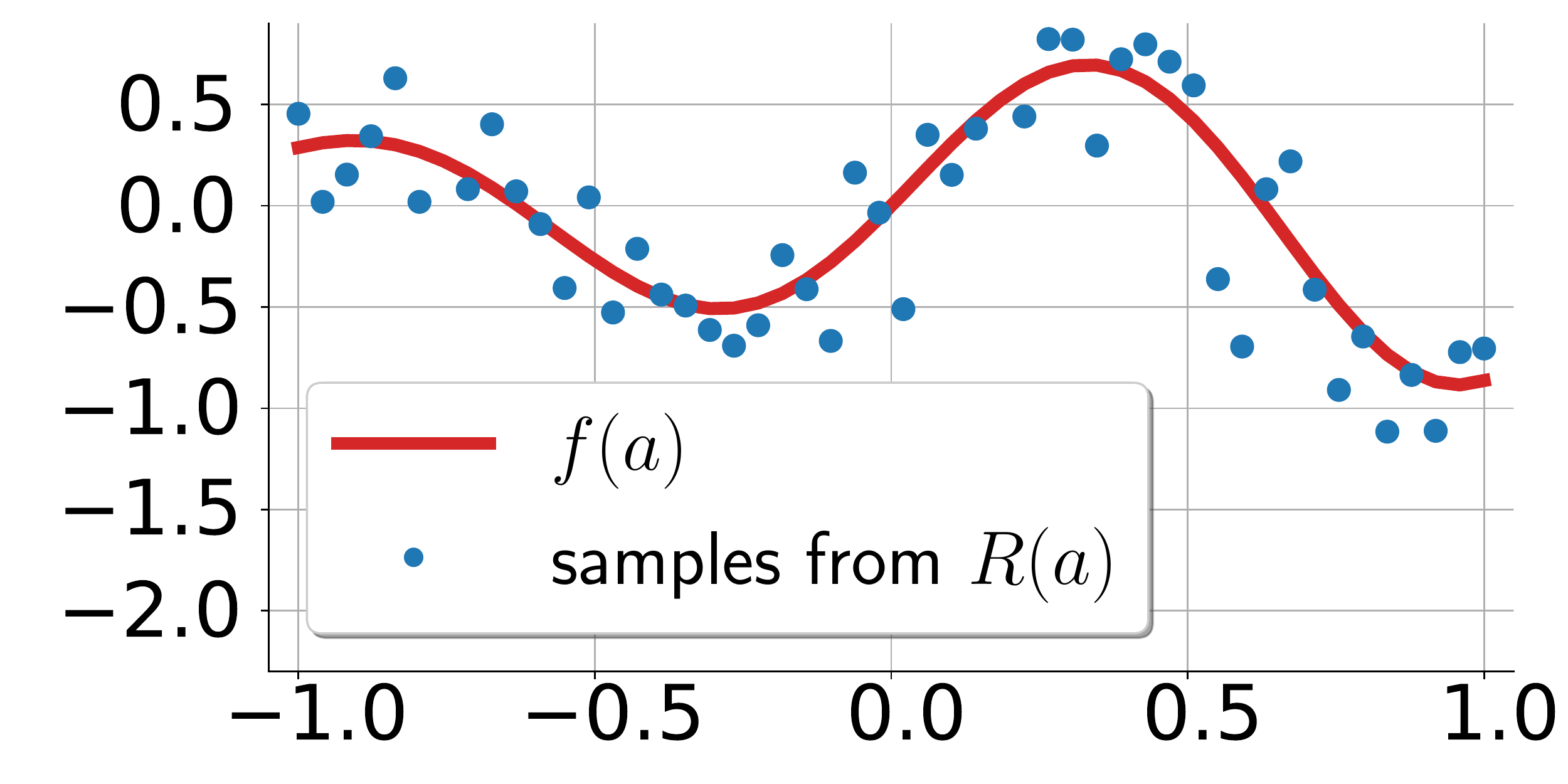}
}\\
\caption{
        Infinite horizon MDP with a single state and one-dimensional action space $[-1, 1]$.
        At each step agent receives a stochastic reward $R(a) \sim \mathcal{N}(f(a), \sigma)$ (see  Appendix~\ref{ap:toy} for details).
}
    \label{fig:mdp}
\end{figure}

\begin{figure}[t]
\includegraphics[width=\columnwidth]{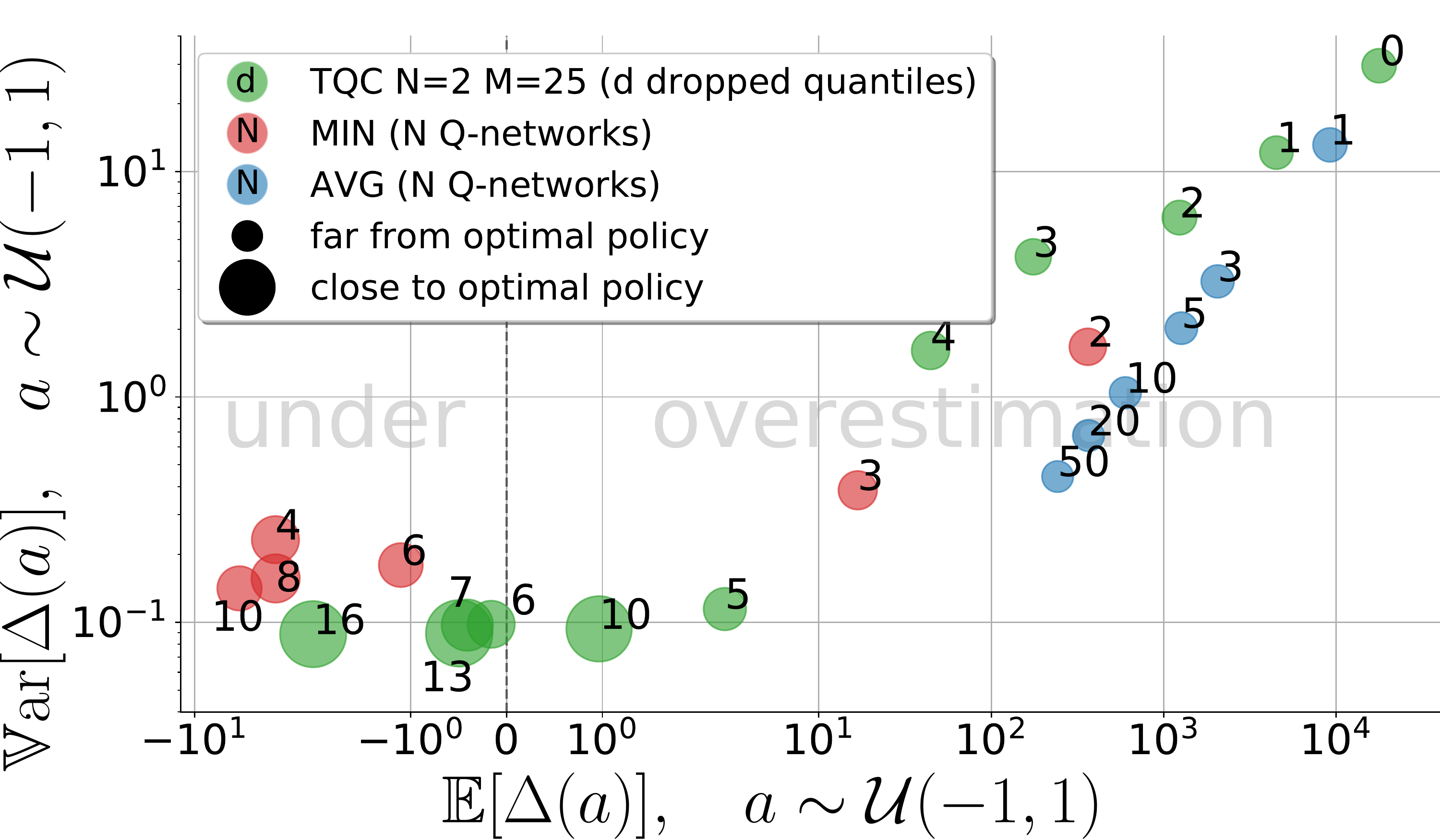}
\caption{
    Robust average ($10\%$ of each tail is truncated) of bias and variance of Q-function approximation for different methods: \textsc{tqc, min, avg}. 
    See the text for the details about axis labels.
}
\label{fig:toy_exp_results}
\end{figure}

In this experiment we evaluate bias correction techniques (Table~\ref{tb:toy_exp_results}) in a single state continuous action infinite horizon MDP (Figure~\ref{fig:mdp}). 
We train Q-networks (or Z-networks, depending on the method) with two hidden layers of size $50$ from scratch on the replay buffer of size $50$ for $3000$ iterations, which is enough for all methods to converge.
We populate the buffer by sampling a reward once for each action from a uniform action grid. 
At each step of temporal difference learning, we use a policy, which is greedy with respect to the objective in Table~\ref{fig:toy_exp_results}.

We define $\Delta(a) := \widehat Q^\pi(a) - Q^\pi(a) $ as a signed discrepancy between the approximate and the true Q-value. 
For \textsc{tqc} $\widehat Q^\pi(a) = \mathbb{E} \widehat Z^\pi(a) = \frac{1}{kN}\sum_{i=1}^{kN} z_{(i)}(a)$.
We vary the parameters controlling the overestimation for each method and report the robust average ($10\%$ of each tail is truncated) over $100$ seeds of 
$\Ep{a \sim \mathcal{U}(-1, 1)} {\Delta(a)}$ and $\Vp{a \sim \mathcal{U}(-1, 1)} {\Delta(a)}$.

For \textsc{avg} and  \textsc{min} we vary the number of networks $N$, for  \textsc{tqc}---the number of dropped quantiles per network $d = M-k$.
We present the results in Figure \ref{fig:toy_exp_results} with bubbles of diameter, inversely proportional to the averaged over the seeds absolute distance between the optimal $a^*$ and the $\argmax$ of the policy objective.

\newcommand{\agg}{\mathrm{aggr}}

\begin{table}[t]
\begin{sc}
    \small
    \renewcommand{\arraystretch}{1.4}
    \caption{
    Bias correction methods. 
    For simplicity, we omit the state $s_0$ from all arguments. 
    }
    \begin{tabular}{l@{\hspace{7pt}}l@{\hspace{7pt}}l}
        \toprule
            \multirow{2}{*}{Method} & Critic target  & \multirow{2}{*}{Policy objective}  \\
            & $r(a) + \gamma \langle \widehat{Q}\text{ or }\widehat{Z} \rangle (a')$& \\
            \midrule
             
            avg & $\widehat{Q}(\cdot) = \frac{1}{N} \sum_{i=1}^N Q_i (\cdot)$ & $\frac{1}{N} \sum_{i=1}^N Q_i (a)$ \\
            
            min & $\widehat{Q}(\cdot) =  \min_i Q_i (\cdot)$ & $\min_i Q_i (a)$ \\ 
            
            tqc & $\widehat{Z}(\cdot) = \frac{1}{kN}\sum_{i=1}^{kN} \delta \left( z_{(i)}(\cdot) \right) $ & $\frac{1}{NM} \sum_{i=1}^{NM} z_{(i)}(a)$ \\
            
        \bottomrule
    \end{tabular}
\label{tb:toy_exp_results}
\end{sc}
\end{table}

The results (Figure~\ref{fig:toy_exp_results}) suggest TQC can achieve the lowest variance and the smallest bias of Q-function approximation among all the competitors. 
The variance and the bias correlate well with the policy performance, suggesting TQC may be useful in practice.

\subsection{Comparative Evaluation}

\begin{figure*}[t]
\includegraphics[width=\textwidth]{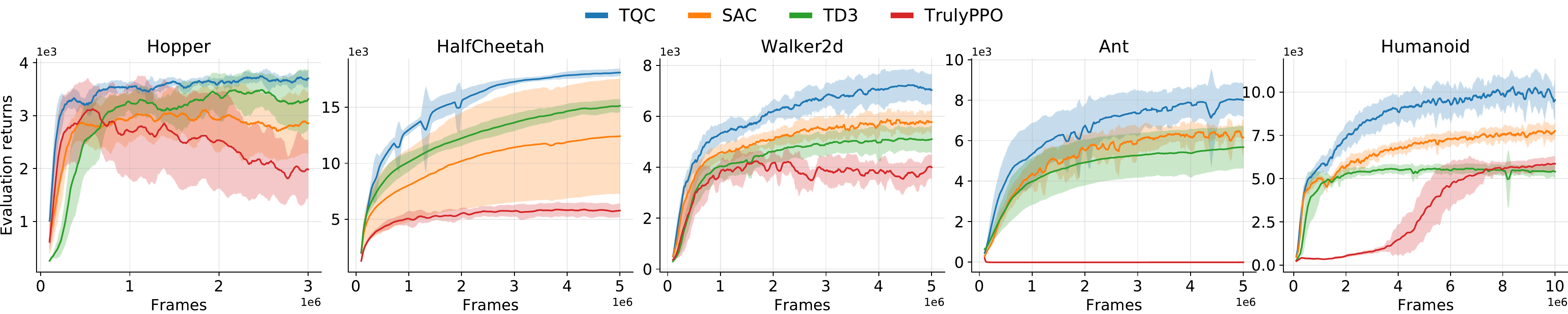}
\caption{
Average performances of methods on MuJoCo Gym Environments with $\pm$ std shaded. 
Smoothed with a window of 100. 
}
\label{fig:main_exps}
\vspace{-.2cm}
\end{figure*}

We compare our method with original implementations of state of the art algorithms: 
SAC\footnote{\url{https://github.com/rail-berkeley/softlearning}}, TrulyPPO\footnote{\url{https://github.com/wangyuhuix/TrulyPPO}}, 
and  TD3\footnote{\url{https://github.com/sfujim/TD3}}.
For HalfCheetah, Walker, and Ant we evaluate methods on the extended frame range: until all methods plateaus (\num{5e6} versus usual \num{3e6}). For Hopper, we extended the range to \num{3e6} steps.

For our method we selected the number of dropped atoms $d$ for each environment independently, based on separate evaluation. Best value for Hopper is $d=5$, for HalfCheetah $d=0$ and for the rest $d=2$.

Figure \ref{fig:main_exps} shows the learning curves. 
In Table~\ref{tb:aggr_methods} we report the average and std of 10 seeds. 
Each seed performance is an average of 100 last evaluations. 
We evaluate the performance every 1000 frames as an average of 10 deterministic rollouts. 
As our results suggest, TQC performs consistently better than any of the competitors.
TQC also improves upon the maximal published score on four out of five environments (Table \ref{tb:max_score}).

\begin{table}[t] 
\caption{
Average and std of the seed returns (thousands). 
The best average return is bolded, and marked with\;$*$ if it is the best at level $0.05$ according to the two-sided Welch's t-test with  Bonferroni correction for  multiple comparison testing. 
}
    \begin{small}
        \begin{sc}
            \begin{tabular}{l@{\hspace{7pt}}c@{\hspace{7pt}}c@{\hspace{7pt}}c@{\hspace{7pt}}c}
                \toprule
Env & TrulyPPO & TD3 & SAC & TQC \\
\midrule
Hop & $1.98(.54)$ & $3.31(.55)$ & $2.86(.58)$ & $\mathbf{3.71(.16)}$\\
HC & $5.78(.62)$ & $15.12(.59)$ & $12.41(5.14)$ & $\mathbf{18.09(.34)}$*\\
Wal & $4.00(.50)$ & $5.11(.52)$ & $5.76(.46)$ & $\mathbf{7.03(.62)}$*\\
Ant & $-0.01(.00)$ & $5.68(1.04)$ & $6.16(.93)$ & $\mathbf{8.01(.87)}$*\\
Hum & $5.86(.45)$ & $5.40(.36)$ & $7.76(.46)$ & $\mathbf{9.54(1.18)}$*\\
                \bottomrule
            \end{tabular}
        \end{sc}
    \end{small}
\label{tb:aggr_methods}    
\end{table}

\begin{table}[t] 
\caption{
    Maximum immediate evaluation score (thousands). 
    Maximum was taken over the learning progress and over 10 seeds (see Figure \ref{fig:main_exps} for the mean plot).
    ARS results were taken from \cite{mania2018simple}.
    The best return per row is bolded. 
}
    \begin{center}
        \begin{small}
            \begin{sc}
                \begin{tabular}{lccc}
                    \toprule
                        Env & ARS-V2-t & SAC & TQC \\
                        \midrule
                        Hop & $3.909$             & $4.232 $  &  $\mathbf{4.288}$  \\
                        HC  & $6.722$             & $16.934$  &  $\mathbf{18.908}$ \\
                        Wal & $\mathbf{11.389}$   & $6.900 $  &  $8.646$ \\
                        Ant & $5.146$             & $7.417 $  &  $\mathbf{9.011}$\\
                        Hum & $11.600$            & $9.411 $  &  $\mathbf{13.163}$ \\
                    \bottomrule
                \end{tabular}
            \end{sc}
        \end{small}
    \end{center}
\label{tb:max_score}    
\end{table}

\section{Ablation study}
\label{sec:ablation}

We ablate TQC on the Humanoid 3D environment, which has the highest resolution power due to its difficulty, and Walker2d---a 2D environment with the largest sizes of action and observation spaces. 
In this section and in the Appendix~\ref{ap:additional_ablation} we average metrics over four seeds.

\subsection{Design choices evaluation}

The path from SAC \cite{haarnoja2018applications} to TQC comprises five modifications: 
Q-network size increase (\textbf{B}ig), quantile Q-network introduction (\textbf{Q}uantile), target distribution truncation (\textbf{T}runcate),  atom pooling (\textbf{P}ool), and ensembling. 
To reveal the effects behind these modifications, we build four methods -- the intermediate steps on the incremental path from SAC to TQC.
Each subsequent method adds the next modification from the list to the previous method or changes the order of applying modifications.
For all modifications, except the final (ensembling), we use $N=2$ networks. 
In all truncation operations we drop $dN$ atoms in total, where $d=2$.

\textbf{B-SAC} is SAC with an increased size of Q-networks (Big SAC):  3 layers with 512 neurons versus 2 layers of 256 neurons in SAC. Policy network size does not change.

\textbf{QB-SAC} is B-SAC with Quantile distributional networks \cite{dabney2018distributional}.
This modification changes the form of Q-networks and the loss function, quantile Huber (\eqref{eq:quant_huber}).
We adapt the clipped double estimator \cite{fujimoto18a} to quantile networks: 
we recover Q-values from distributions and use atoms of the argmin of Q-values to compute the target distribution
$ Y^{QB}(s,a) := \frac{1}{M} \sum_{m=1}^{M} \delta \left( y^m (s, a) \right)$, where $y^m(s,a)$ is 
\begin{equation}
r(s,a) +  \gamma [\theta^{m}_{\overline{\psi}_{j(s',a')}}(s', a')  - \alpha \log \pi_\phi(a' | s')] 
\end{equation}
and $j(s', a') := \argmin_n \frac{1}{M} \sum_{m=1}^{M} \theta_{\overline{\psi}_n}^{m}(s', a')$.

\textbf{TQB-SAC} is QB-SAC with individual truncation instead of $\min$: $Z_{\psi_n}$ is trained to approximate the truncated temporal difference distribution $Y_n^{TQB}(s,a)$, which is based on the predictions of the single target network $Z_{\overline{\psi}_n}$ only. 
That is, $Z_{\psi_n}$ is trained to approximate 
$Y_n^{TQB}(s,a) := \frac{1}{k} \sum_{m=1}^{k} \delta ( y^m_n (s, a) )$, where $y^m_n (s, a)$ is 

\begin{equation}
  r(s,a) + 
  \gamma [\theta^{m}_{\overline{\psi}_n}(s', a')  - \alpha \log \pi_\phi(a' | s')]. 
\end{equation}

\textbf{PTQB-SAC} is TQB-SAC with pooling: $Z_{\psi_n}$ approximates the mixture of (already truncated) $Y_n^{TQB}(s,a), n \in [1..N]$:
\begin{equation}
Y^{PTQB}(s,a) := \frac{1}{kN} \sum_{n=1}^{N}\sum_{m=1}^{k} \delta ( y^m_n (s, a) ).
\end{equation}
\textbf{TQC = TPQB-SAC} is PTQB-SAC with pooling and truncation operations swapped. This modification drops the same number of atoms as two previous methods, but differs in \textit{which} atoms are dropped. 
TQC drops $d N$ largest \textit{from the union of} $N$ critics predictions. 
While each of PTQB-SAC and TQB-SAC (no pooling) drops $d$ largest atoms \textit{from each} of $N$ critics.

\textbf{Ensembling} To illustrate the advantage brought by ensembling, we include the results for TQC with two and five Z-networks.

\begin{figure}[t]
\includegraphics[width=\columnwidth]{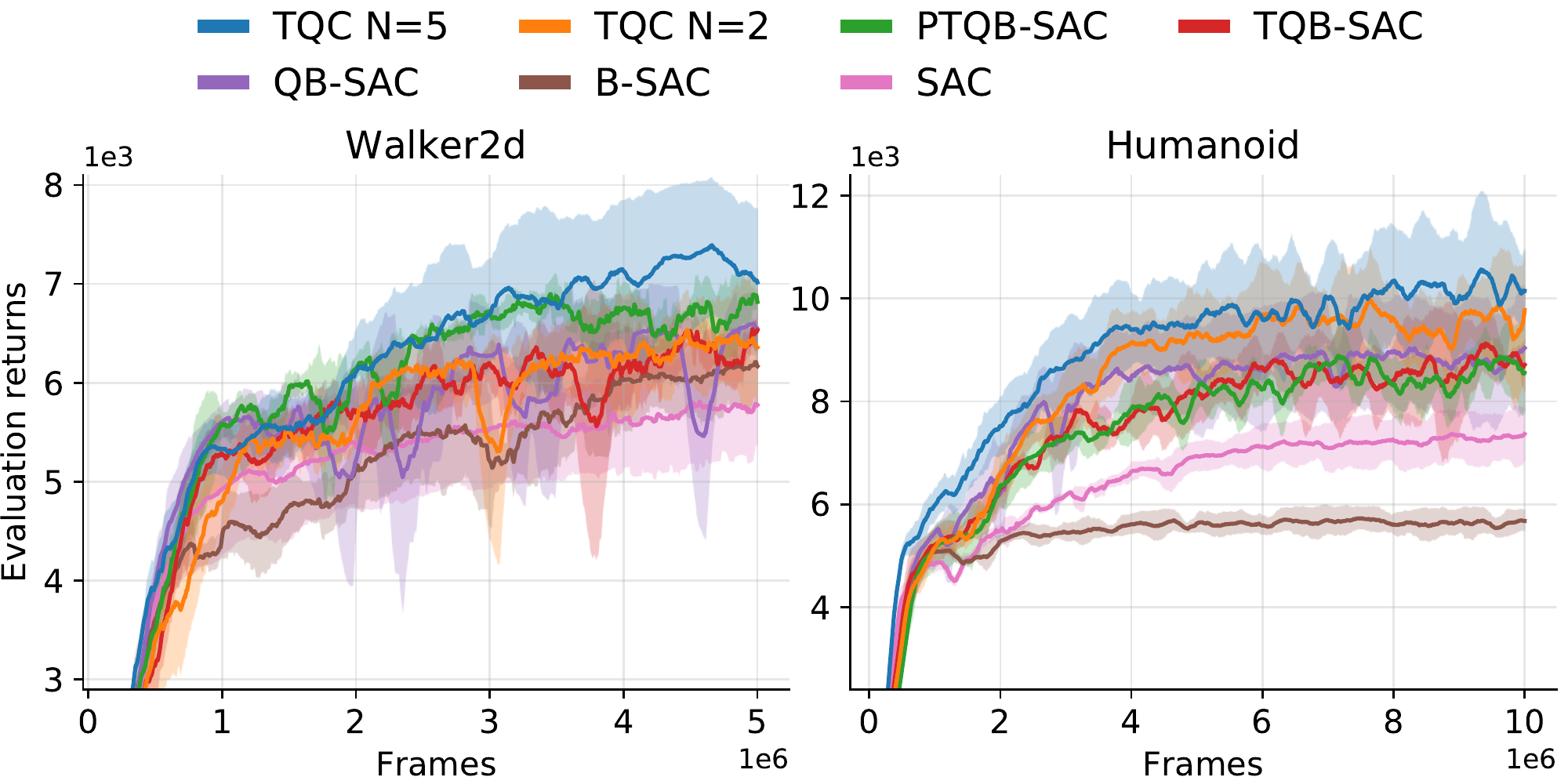}
\caption{
Design choices evaluation. $N=2$ where isn't stated otherwise, $d=2$ and $M=25$ where applicable. 
Smoothed with a window of 200, $\pm$ std is shaded. 
}
\label{ablation_exps}
\vspace{-.2cm}
\end{figure}

The ablation results (Figure~\ref{ablation_exps}) suggest the following.  
The increased network size does not necessarily improve SAC (though some improvement is visible on Walker2d).
The quantile representation improves the performance in both environments with the most notable improvement on Humanoid. 
Among the following three modifications---individual truncation, and two orders of applying the pooling and truncation---the TQC is a winner for Humanoid and 
"truncate-pooling" modification seems to be better on Walker2d.
Overall, truncation stabilizes results on Walker2d and seems to reduce the seed variance on Humanoid. 
Finally, the ensembling consistently improves results on both environments.

\subsection{Sensitivity to hyperparameters}
\label{sec:hyperparms_sens}

\begin{figure}[t]

\includegraphics[width=\columnwidth]{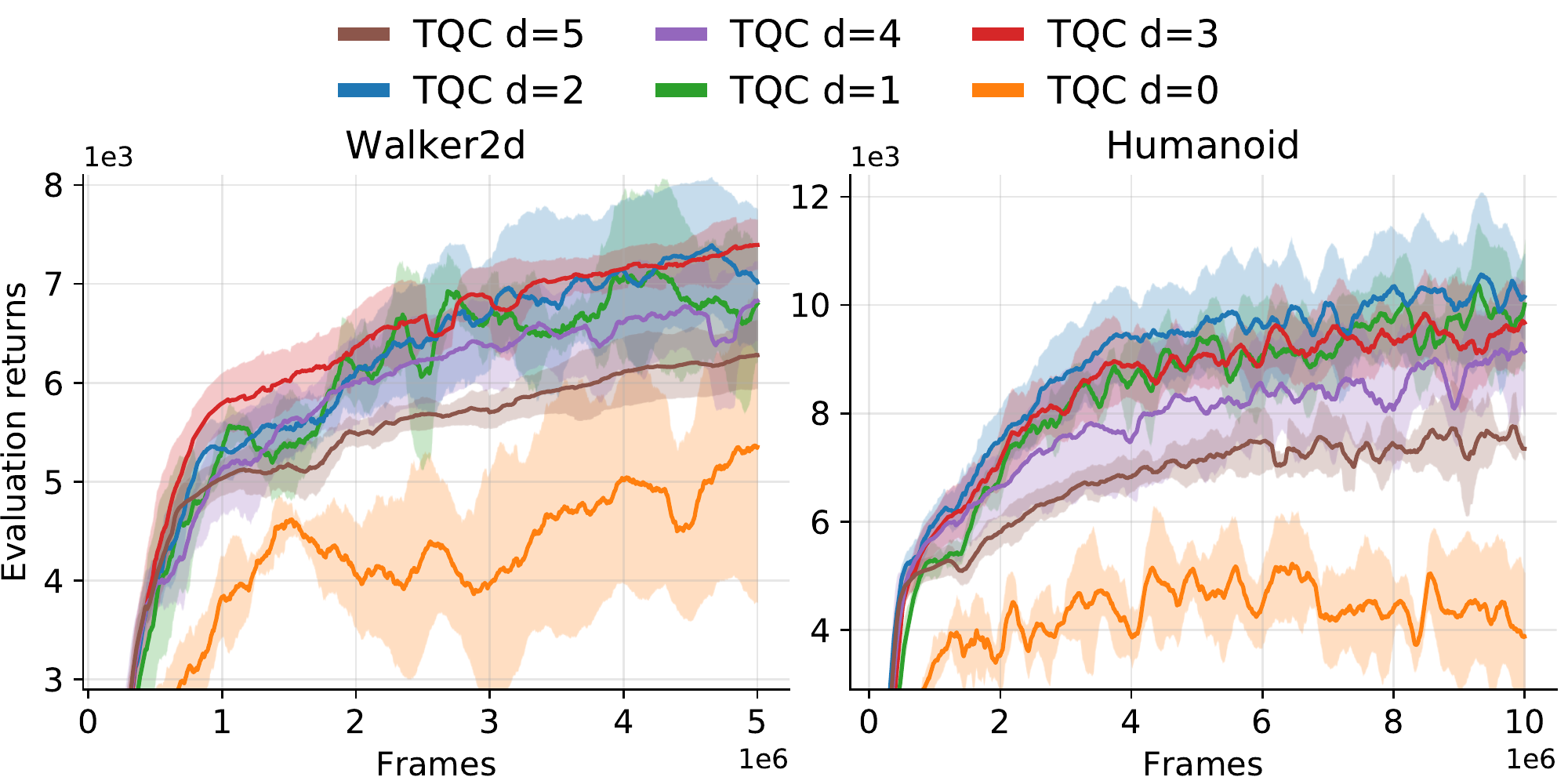}
\caption{
Varying the number of dropped atoms per critic $d$.  $N=5$ networks, $M=25$ atoms. 
Smoothed with a window of 200, $\pm$ std is plotted. 
}
\label{fig:dropped_quantiles}
\end{figure}

\textbf{Number of truncated quantiles}~~
In this experiment we vary the number of atoms (per network) to drop in the range $d \in [0..5]$. 
The total number of atoms dropped is $dN$.
We fix the number of atoms for each Q-network to $M=25$. 
The results (Figure~\ref{fig:dropped_quantiles}) show that (1) truncation is essential and (2) there is an optimal number of dropped atoms (i.e., $d=2$ or $d=3$).

\textbf{Number of total quantiles}~~
In this experiment we vary the total number of atoms $M \in \{10, 15, 25, 35, 50\}$ and adjust the number of dropped quantiles to keep the truncation ratio approximately constant. 
The results (Appendix~\ref{ap:additional_ablation}) suggest this parameter does not have much influence, except for the case of very small $M$, such as $10$. 
For $M\geq15$ learning curves are indistinguishable.

\textbf{Number of Z-networks}~~
In this experiment we vary the number of Z-networks $N \in \{1, 2, 3, 5, 10\}$. 
The results (Appendix~\ref{ap:additional_ablation}) suggest that (1) a single network is consistently inferior to larger ensembles and (2) performance improvement saturates at approximately $N=3$.  

\label{sec:limitations}
\begin{table}[t]
\caption{Time measurements (in seconds) of a single training epoch (1000 frames), averaged over 1000 epochs, executed on the Tesla P40 GPU.}
    \label{tb:time}
    \vskip 0.15in
    \begin{center}
        \begin{small}
            \begin{sc}
                \begin{tabular}{ccccc}
                    \toprule
                        Env & SAC & B-SAC & TQC N=2 & TQC N=5 \\
                        \midrule
                        Walker2d & 9.5 & 13.9 & 14.1 & 32.4 \\
                        Humanoid & 10.7 & 16.5 & 17.4 & 36.8 \\
                    \bottomrule
                \end{tabular}
            \end{sc}
        \end{small}
    \end{center}
    \vskip -0.1in
\end{table}
Ensembling and distributional networks incur additional computational overhead, which we quantify for different methods in Table \ref{tb:time}.

\section{Related work}
\paragraph{Distributional perspective}
Since the introduction of distributional paradigm (see \citet{white1988mean} and references therein) and its reincarnation for deep reinforcement learning \cite{bellemare2017distributional} a great body of research emerged.
\citeauthor{dabney2018distributional} proposed a method to learn quantile values (or locations) for a 
uniform grid of fixed \cite{dabney2018distributional} or sampled \cite{dabney2018implicit} quantile fractions. 
\citet{yang2019fully}  proposed a method to learn both quantile fractions and quantile values (i.e. locations and probabilities of elements in the mixture approximating the unknown distribution). 
\citet{choi2019distributional} used a mixture of Gaussians for approximating the distribution of returns. 
Most of these works, as well as their influential follow-ups,  such as \cite{hessel2018rainbow}, are devoted to the discrete control setting. 

The adoption of the distributional paradigm in continuous control, to the best of our knowledge, starts from D4PG \cite{barth2018distributed}---a distributed distributional off-policy algorithm building on the C51 \cite{bellemare2017distributional}.
Recently, the distributional paradigm was adopted in distributed continuous control for robotic grasping \cite{bodnar2019quantile}. 
The authors proposed Q2-Opt as a collective name for two variants: based on QR-DQN \cite{dabney2018distributional}, and on IQN \cite{dabney2018implicit}. 
In contrast to D4PG and Q2-Opt, we focus on the usual, non-distributed setting and modify the target on which the critic is trained. 

A number of works develop exploration methods based on the quantile form of value-function.
DLTV \cite{mavrin2019distributional} uses variability of the quantile distribution in the exploration bonus. 
QUOTA \cite{zhang2019quota}---the option-based modification of QR-DQN---partitions a range of quantiles into contiguous windows and trains a separate intra-option policy to maximize an average of quantiles in each of the windows.
Our work proposes an alternative method for critic training, which is unrelated to the exploration problem.

Most importantly, our work differs from the research outlined above in our aim to control the \textit{overestimation bias} by leveraging quantile representation of a critic network.

\paragraph{Overestimation bias} The overestimation bias is a long-standing topic in several research areas. It is known as the max-operator bias in statistics \cite{d2017estimating} and as  the "winner's curse" in economics  \cite{smith2006optimizer,thaler2012winner}.

\textit{The statistical community} studies estimators of the maximum expected value of a set of independent random variables. 
The simplest estimator---the maximum over sample means,  Maximum Estimator (ME)---is biased positively, while for many distributions, such as Gaussian, an unbiased estimator does not exist \cite{ishwaei1985non}. 
The Double Estimator (DE) \cite{stone1974cross, van2013estimating} uses cross-validation to decorrelate the estimation of the argmax and of the value for that argmax.
\citet{he2019interleaved} proposed a coupled estimator as an extension of DE to the case of partially overlapping cross-validation folds. 
Many works have aimed at alleviating the negative bias of DE, which in absolute value can be even larger than that of ME.
\citet{discrete2016estimating} assumed the Gaussian distribution for the sample mean and proposed Weighted Estimator (WE) with a bias in between of that for ME and DE. \citet{imagaw2017estimating} improved WE by using UCB for weights computation.
\citet{d2017estimating} assumed a certain spatial correlation and extended WE to continuous sets of random variables.
The problem of overestimation has also been discussed in the context of optimal stopping and sequential testing \cite{kaufmann2018sequential}.

\textit{The reinforcement learning community} became interested in the bias since the work of \citet{thrun1993issues}, who attributed a  systematic overestimation to the generalization error. 
The authors proposed multiple ways to alleviate the problem, including (1) bias compensation with additive pseudo costs and (2)  underestimation (e.g., in the uncertain areas).   
The underestimation concept became much more prominent, while the adoption of "additive compensation" is quite limited to date  \cite{patnaik2008q, lee2012intelligent}.

\citet{hasselt2010double} proposed  Double Estimation in Q-learning, which was subsequently adapted to neural networks as Double DQN \citet{hasselt2015double}.
Subsequently, \cite{zhang2017weighted} and \citet{lv2019stochdoubledqn} introduced the Weighted Estimator in the reinforcement learning community.

Another approach against overestimation and overall Q-function quality improvement is based on the idea of averaging or ensembling. 
Embodiments of this approach are based on  
dropout \citet{anschel2017averaged},
employing previous Q-function approximations \cite{azar2011speedy, anschel2017averaged},  
the linear combination between $\min$ and $\max$ over the pool of Q-networks \cite{li2019mixing, kumar2019stabilizing},
or the random mixture of predictions from the pool \cite{agarwal2019striving}.
\citet{buckman2018sample} reported the reduction in overestimation originating from ensembling in model-based learning.

In \textit{continuous} control, \citet{fujimoto18a} proposed the TD3 algorithm, taking the $\min$ over two approximators of Q-function to reduce the overestimation bias.
Later, for \textit{discrete} control \citet{lan2020maxmin} developed a MaxMin Q-learning, taking the $\min$ over more than two Q-functions.
We build upon the minimization idea of \citet{fujimoto18a} and, following \citet{lan2020maxmin} use multiple approximators. 

Our work differs in that we do not propose to control the bias by choosing \textit{between} multiple approximators or \textit{weighting} them.
For the first time, we propose to successfully control the overestimation even for a single approximator and use ensembling only to improve the performance further.

\section{Conclusion and Future Work}

In this work, we propose to control the overestimation bias on the basis of aleatoric uncertainty. 
The method we propose comprises three essential ideas: distributional representations, truncation of a distribution, and ensembling. 

Simulations reveal favorable properties of our method: low expected variance of the approximation error as well as the fine control over the under- and overestimation.
The exceptional results on the standard continuous control benchmark suggest that distributional representations may be useful for controlling the overestimation bias. 

Since little is known about the connection between aleatoric uncertainty and overestimation, we see the  investigation of it as an exciting avenue for future work.

\section{Acknowledgements}

We would like to thank Artem Sobolev, Arsenii Ashukha, Oleg Ivanov and Dmitry Nikulin for their comments and suggestions regarding the early versions of the manuscript. 
We also thank the anonymous reviewers for their feedback. 

This research was supported in part by the Russian Science Foundation grant no.~19-71-30020.

\bibliography{example_paper}
\bibliographystyle{icml2020}

\appendix

\onecolumn
\section{Experimental setting}
\label{ap:exp_setting}

We would like to caution about the use of MuJoCo 2.0 with versions of Gym at least up to \texttt{v0.15.4} (the last released at the moment). For these versions Gym incorrectly nullifies state components corresponding to contact forces, which, in turn makes results incomparable to previous works. 

In our work we use MuJoCo 1.5 and  \verb|v3| versions of environments. Versions of all other packages we used are listed in the Conda environment file, distributed with the source code\footnote{See the code attached.}.

\section{Hyperparameters}
\label{ap:hyperparams}

Critic networks are fully-connected, with the last layer output size equal to the number of atoms $M$.
\begin{table}[h]
\caption{Hyperparameters values.}
\label{time-measurements}
\vskip 0.15in
\begin{center}
\begin{small}
\begin{sc}
\begin{tabular}{lcc}
\toprule
Hyperparameter & TQC & SAC \\
\midrule
Optimizer & \multicolumn{2}{c}{Adam} \\
Learning rate & \multicolumn{2}{c}{\num{3e-4}} \\
Discount $\gamma$ & \multicolumn{2}{c}{0.99} \\
Replay buffer size & \multicolumn{2}{c}{\num{1e6}} \\
Number of critics $N$ & 5 & 2 \\
Number of hidden layers in critic networks & 3 & 2\\
Size of hidden layers in critic networks & 512 & 256 \\
Number of hidden layers in policy network & \multicolumn{2}{c}{2} \\
Size of hidden layers in policy network & \multicolumn{2}{c}{256} \\
Minibatch size & \multicolumn{2}{c}{256} \\
Entropy target $\mathcal{H}_T$ &  \multicolumn{2}{c}{$- \dim \mathcal{A}$} \\
Nonlinearity & \multicolumn{2}{c}{ReLU} \\
Target smoothing coefficient $\beta$ & \multicolumn{2}{c}{0.005} \\
Target update interval & \multicolumn{2}{c}{1} \\
Gradient steps per iteration & \multicolumn{2}{c}{1} \\
Environment steps per iteration & \multicolumn{2}{c}{1} \\
\midrule
Number of atoms $M$ & 25 & ---\\
Huber loss parameter $\kappa$ & 1 & --- \\
\bottomrule
\end{tabular}
\end{sc}
\end{small}
\end{center}
\vskip -0.1in
\end{table}

\begin{table}[h]
\caption{Environment dependent hyperparameters for TQC.}
\label{time-measurements}
\vskip 0.15in
\begin{center}
\begin{small}
\begin{sc}
\begin{tabular}{lcc}
\toprule
Environment & Number of dropped atoms, $d$ & Number of environment steps \\
\midrule
Hopper & 5 & \num{3e6} \\
HalfCheetah & 0 & \num{5e6} \\
Walker2d & 2 & \num{5e6} \\
Ant & 2 & \num{5e6} \\
Humanoid & 2 & \num{10e6} \\
\bottomrule
\end{tabular}
\end{sc}
\end{small}
\end{center}
\vskip -0.1in
\end{table}

\section{Toy experiment setting}
\label{ap:toy}      

\begin{figure}[ht]
\begin{center}
\centerline{\includegraphics[width=0.5\columnwidth]{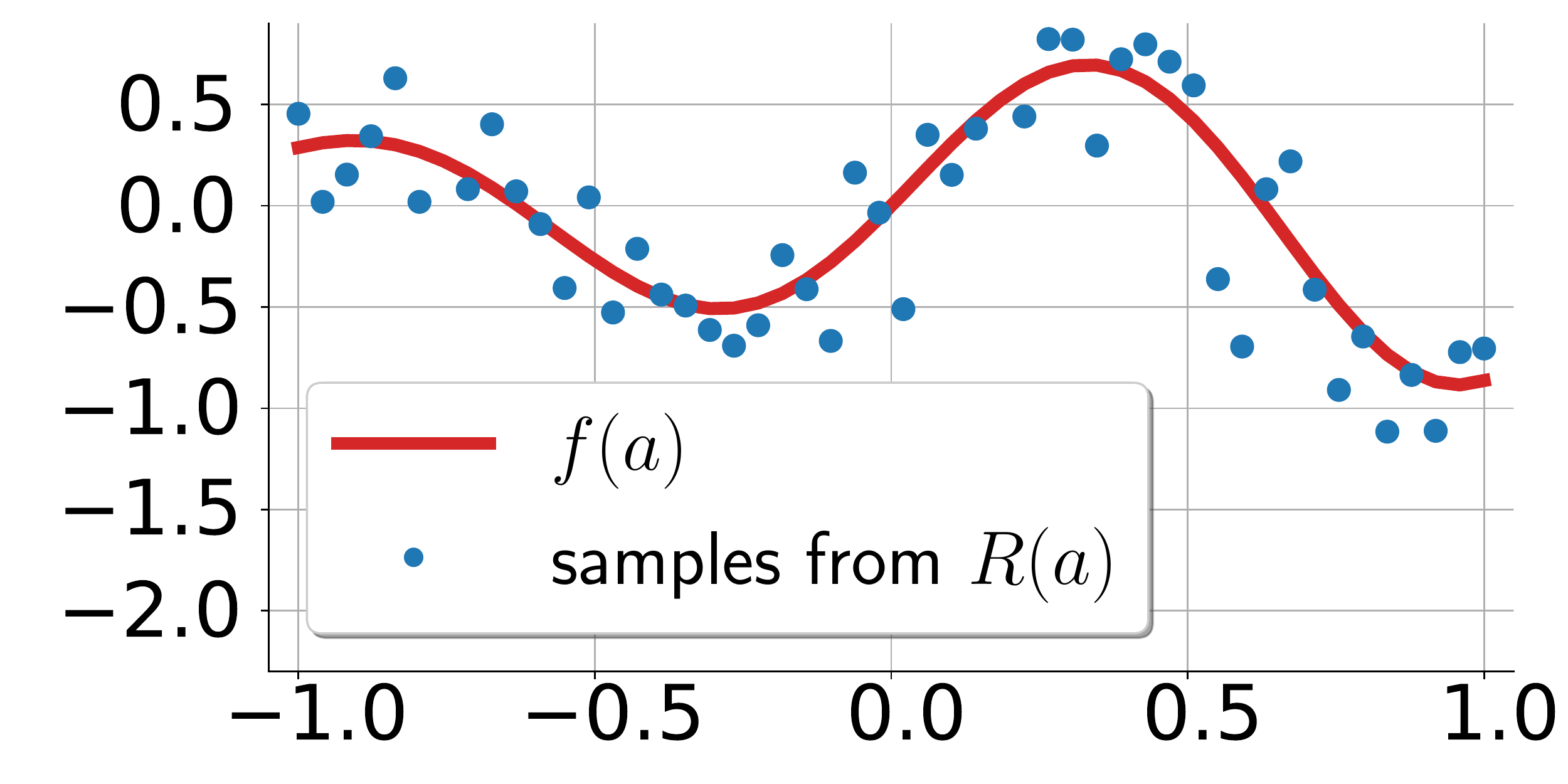}}
\caption{Reward function. 
$x$-axis represents one dimensional action space, 
$y$-axis - corresponding stochastic rewards and their expectation.
}
\label{fig:reward_function}
\end{center}
\end{figure}

The task is simplistic infinite horizon MDP ($\mathcal{S}, \mathcal{A}, \mathcal{P}, \mathcal{R}, p_0$) with only one state $\mathcal{S}=\{s_0\}$ and 1-dimensional action space $\mathcal{A} = [-1, 1]$.
Since there is only one state, the state transition function $\mathcal{P}$ and initial state distribution $p_0$ are delta functions. 
On each step agent get stochastic reward $r(a) \sim f(a) +  \mathcal{N}(0, \sigma)$, where $\sigma=0.25$. 
Mean reward function is the cosine with slowly increasing amplitude (Figure~\ref{fig:reward_function}):
$$
    f(a) = 
    \big[A_0 + \frac{A_1 - A_0}{2}  (a + 1)\big] \cos \nu a,
    \text{ ~~ where }A_0 = 0.3;\quad A_1=0.9;\quad \nu = 5
$$
The discount factor is $\gamma=0.99$.

Such parameters gives raise to three local maxima: near the left end $a\approx -0.94$, in the right half $a^* \approx 0.31$ (global) and at the right end $a = 1$.  Optimal policy in this environment always selects $a^* = \arg \max_a f(a)$.

In the toy experiment we evaluate bias correction techniques (Table \ref{fig:toy_exp_results}) in this MDP. 
We train Q-networks (or Z-networks, depending on the method) with two hidden layers of size $50$ from scratch on the replay buffer of size $50$ for $3000$ iterations.
We populate the buffer by sampling a reward once for each action from a uniform action grid of size $50$. 
At each step of temporal difference learning, we use a policy, which is greedy with respect to the objective in Table~\ref{fig:toy_exp_results}.

We define $\Delta(a) := \widehat Q^\pi(a) - Q^\pi(a) $ as a signed discrepancy between the approximate and the true Q-value. 
For TQC $\widehat Q^\pi(a) = \mathbb{E} \widehat Z^\pi(a) = \frac{1}{kN}\sum_{i=1}^{kN} z_{(i)}(a)$.
We vary the parameters controlling the overestimation for each method and report the robust average ($10\%$ of each tail is truncated) over $100$ seeds of 
$\Ep{a \sim \mathcal{U}(-1, 1)} {\Delta(a)}$ and $\Vp{a \sim \mathcal{U}(-1, 1)} {\Delta(a)}$. Expectation and variance estimated over dense uniform grid of actions of size $2000$ and then averaged over seeds.

For \textsc{avg} and  \textsc{min} we vary the number of networks $N$ from $[3, 5, 10, 20, 50]$ and $[2, 3, 4, 6, 8, 10]$ correspondingly. For \textsc{tqc} --- number of dropped quantiles per network $d = M-k$ from $[0, 1, 2, 3, 4, 5, 6, 7, 10, 13, 16]$ out of $25$.
We present the results in Figure \ref{fig:toy_exp_results} with bubbles of diameter, inversely proportional to the averaged over the seeds absolute distance between the optimal $a^*$ and the $\argmax$ of the policy objective.

To prevent interference of policy optimization subtleties into conclusions about Q-function approximation quality, we use implicit deterministic policy induced by value networks: the argmax of the approximation. 
To find the maximum, we evaluated the approximation over the dense uniform grid in the range $[-1, 1]$ with a step $\Delta a = 0.001$. 

Each dataset consists of uniform grid of actions and sampled corresponding rewards. 
For each method we average results over several datasets and evaluate on different dataset sizes.
In this way current policy defined implicitly as greedy one with respect to value function. 
This policy doesn't interact with the environment instead actions predefined to be uniform.

\section{Mistakenly unreferenced appendix section}
We are sorry for the void section.
We keep this section to make references in the main text valid and will remove it in the camera ready version. 

\newpage

\section{Additional experimental results}
\label{ap:additional_ablation}
\subsection{Number of critics}

\begin{figure*}[h]
\centering
\includegraphics[width=0.85\textwidth]{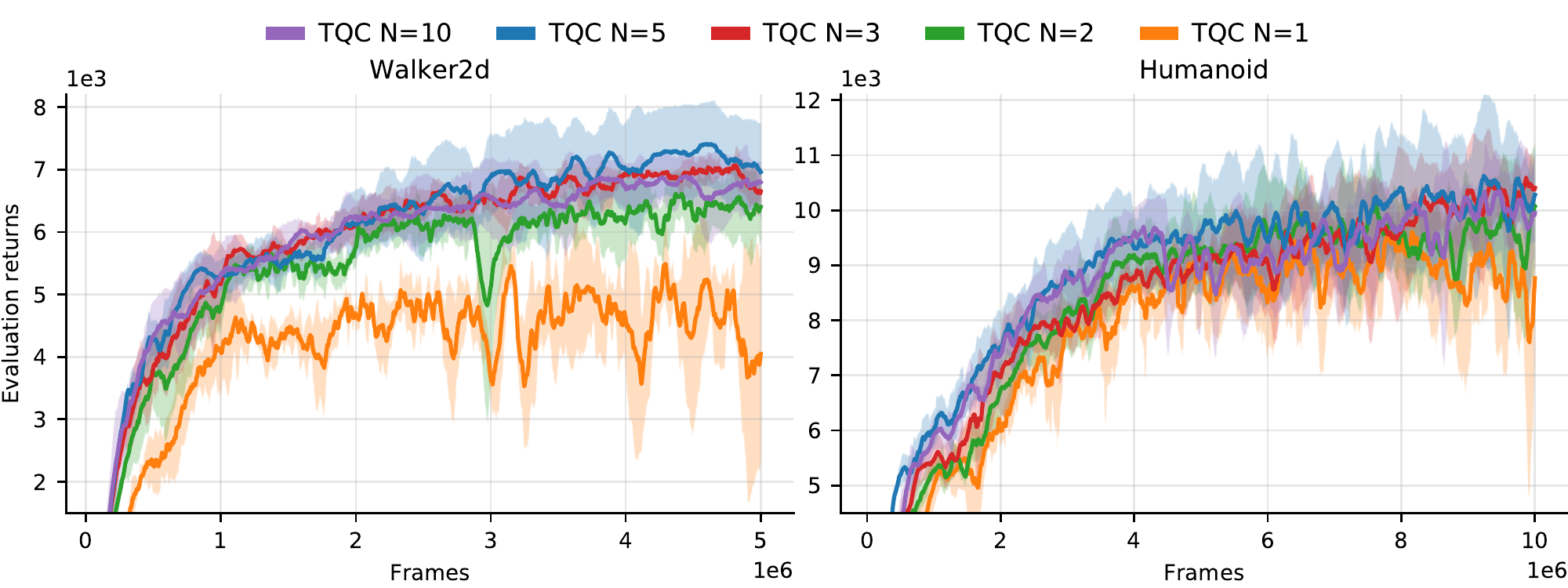}
\caption{
Varying the number of critic networks $N$ for TQC with $M=25$ atoms per critic and $d=2$ of dropped atoms per critic.  
Smoothed with a window of 100, $\pm$ std is plotted.
}
\label{fig:n_critics_tqc}
\end{figure*}

\FloatBarrier
\subsection{Total number of atoms $M$}
\FloatBarrier
\begin{figure}[!h]
\centering
\includegraphics[width=0.85\columnwidth]{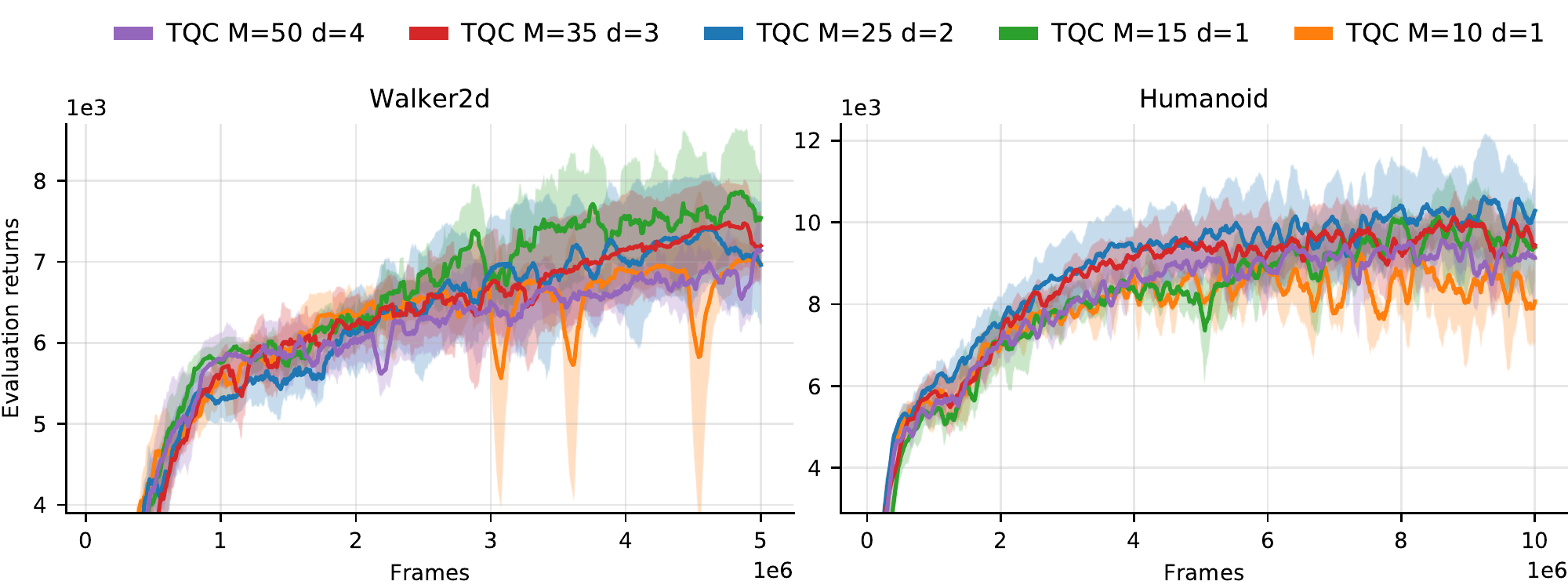}
\label{fig:total_quantiles}
\caption{
Varying the number of atoms per critic $M$ for TQC with $N=5$ critics and $d=2$ dropped atoms per critic.  
Smoothed with a window of 100, $\pm$ std is plotted.
}
\end{figure}
\FloatBarrier
\subsection{Removed atoms stats}

\begin{figure}[h]
\centering
\includegraphics[width=0.85\columnwidth]{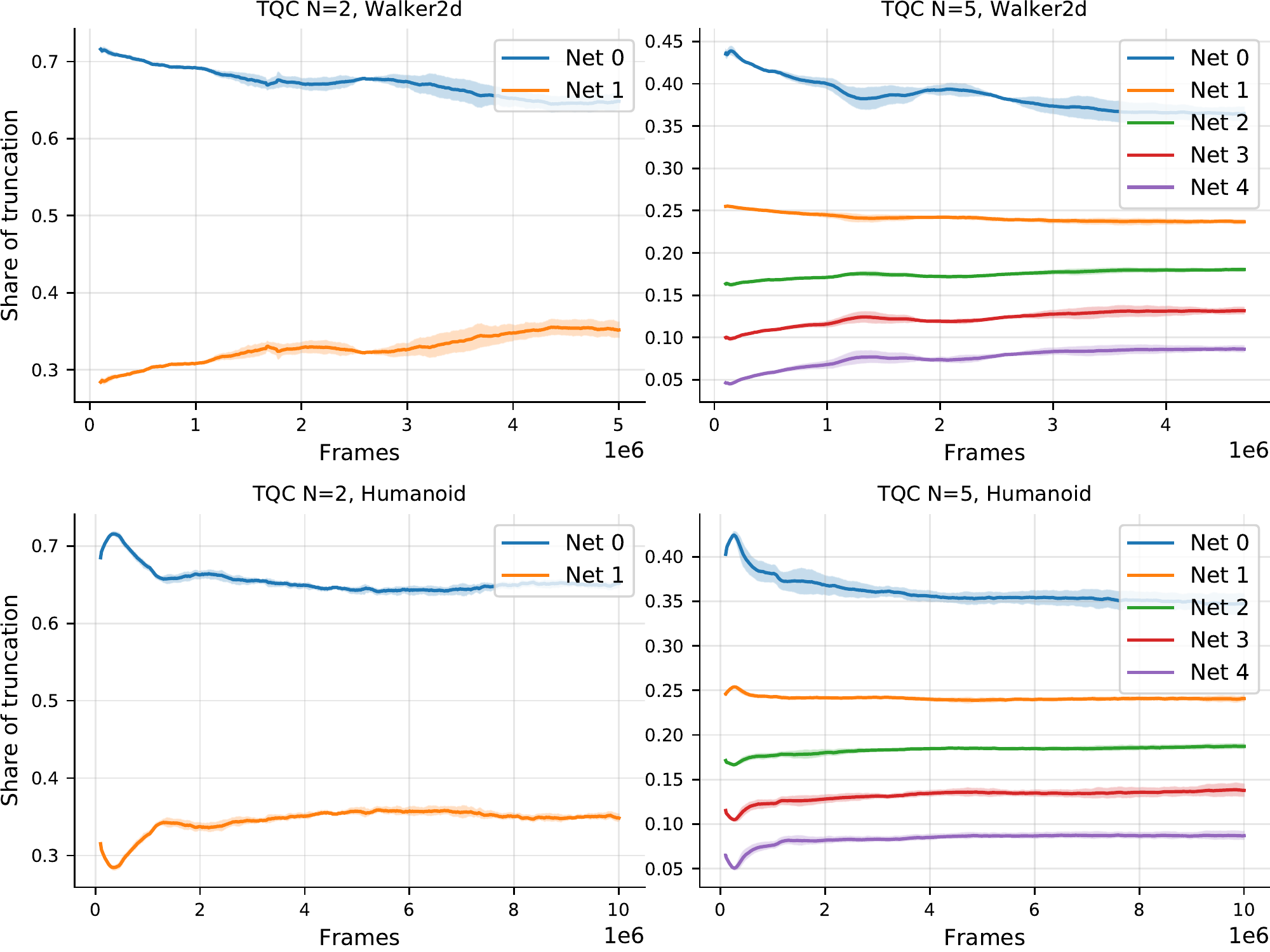}
\caption{
Proportions of atoms dropped per critic, \textit{sorted} and averaged over the minibatches drawn from the experience replay for 
TQC with $N=2$ and $N=5$ critics with $M=25$ and $d=2$ dropped atoms per critic.  
For example, the upper right plot, should be read as "on average the largest proportion of dropped atoms per critic is 35\%, i.e. out of $N \cdot d = 10$ atoms dropped approximately $4$ were predicted by a single critic. 
Smoothed with a window of 100, $\pm$ std is plotted.
}
\label{fig:trunc_share}
\end{figure}

TQC drops atoms with largest locations after the pooling of atoms from multiple Z-networks. 
Experimentally, this procedure drops more atoms for some Z-networks than for the others. 
To quantify this disbalance, we compute the ratio of dropped atoms to the total $M$ atoms for each of Z-networks. 
These proportions, once sorted and averaged over the replay, are approximately constant throughout learning:  $65/35$\% for $N=2$ and  $35/25/18/13/9$\% for $N=5$  (Figure \ref{fig:trunc_share}).

\begin{figure}[ht]
\centering
\includegraphics[width=0.85\columnwidth]{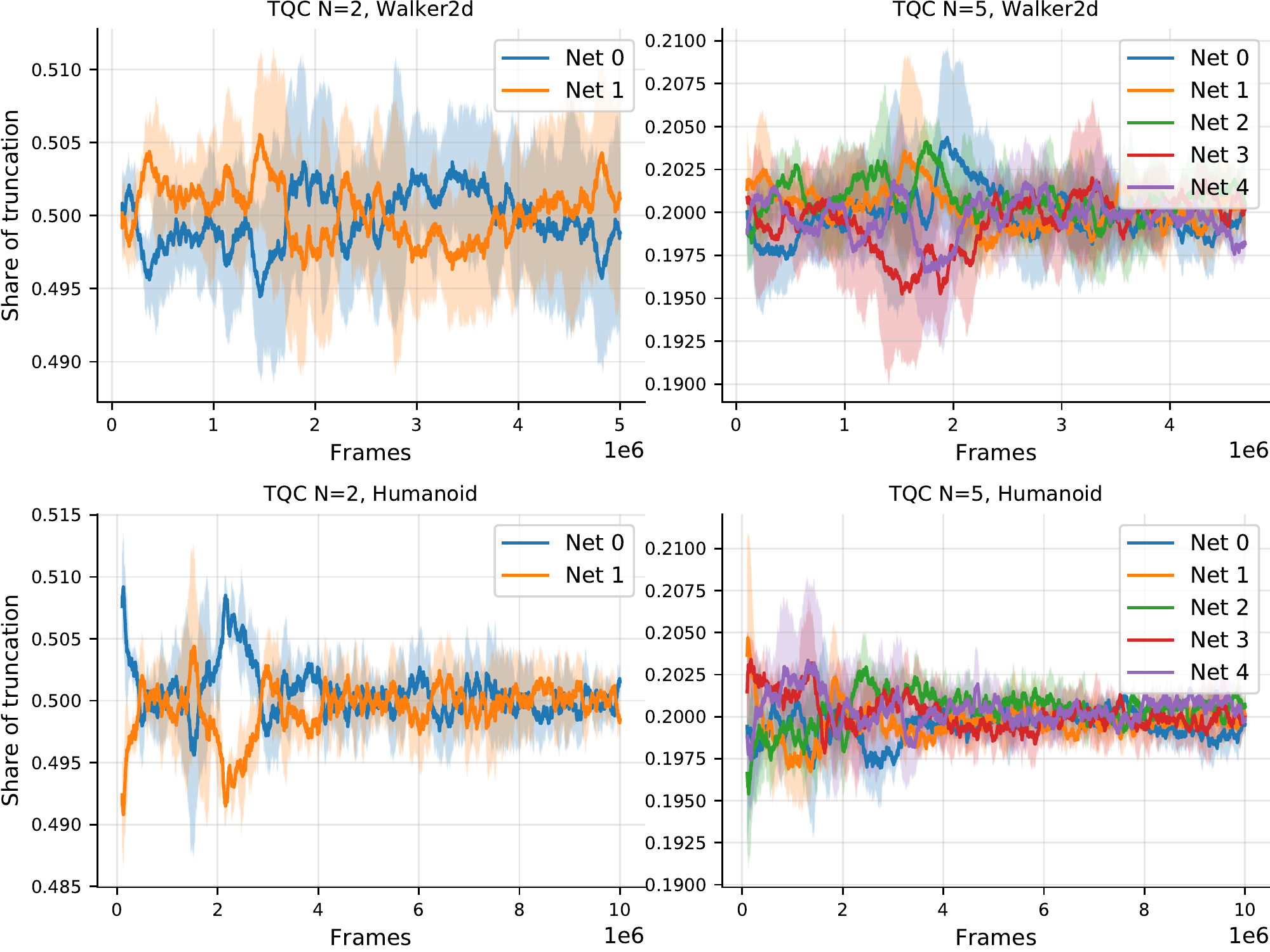}
\caption{
Proportions of atoms dropped per critic, averaged over the minibatches drawn from the experience replay for 
TQC with $N=2$ and $N=5$ critics with $M=25$ and $d=2$ dropped atoms per critic. Same plot as \ref{fig:trunc_share}, but without sorting. 
The figure illustrates that there is no a single critic consistently overestimating more than the others over the whole state action space. 
Smoothed with a window of 100, $\pm$ std is plotted.
}
\label{fig:trunc_share_unsorted}

\end{figure}

Interestingly, without sorting the averaging over the replay gives almost perfectly equal proportions (Figure \ref{fig:trunc_share_unsorted}).
These results suggest, that a critic overestimates in some regions of the state action space more, than any other critic.
In other words, in practice the systematic overestimation of a single critic (w.r.t. other critics predictions) on the whole state action space does not occur.

\FloatBarrier
\clearpage
\subsection{Clipped Double Q-learning}

To ensure that it is not possible to match the performance of TQC with careful tuning of previous methods, we varied the number of critic networks used in the Clipped Double Q-learning estimate \cite{fujimoto18a} for SAC \cite{haarnoja2018applications}. 
The larger the number of networks under the $\min$, the more the underestimation \cite{lan2020maxmin}.

\begin{figure*}[ht]
\centering
\includegraphics[width=0.8\textwidth]{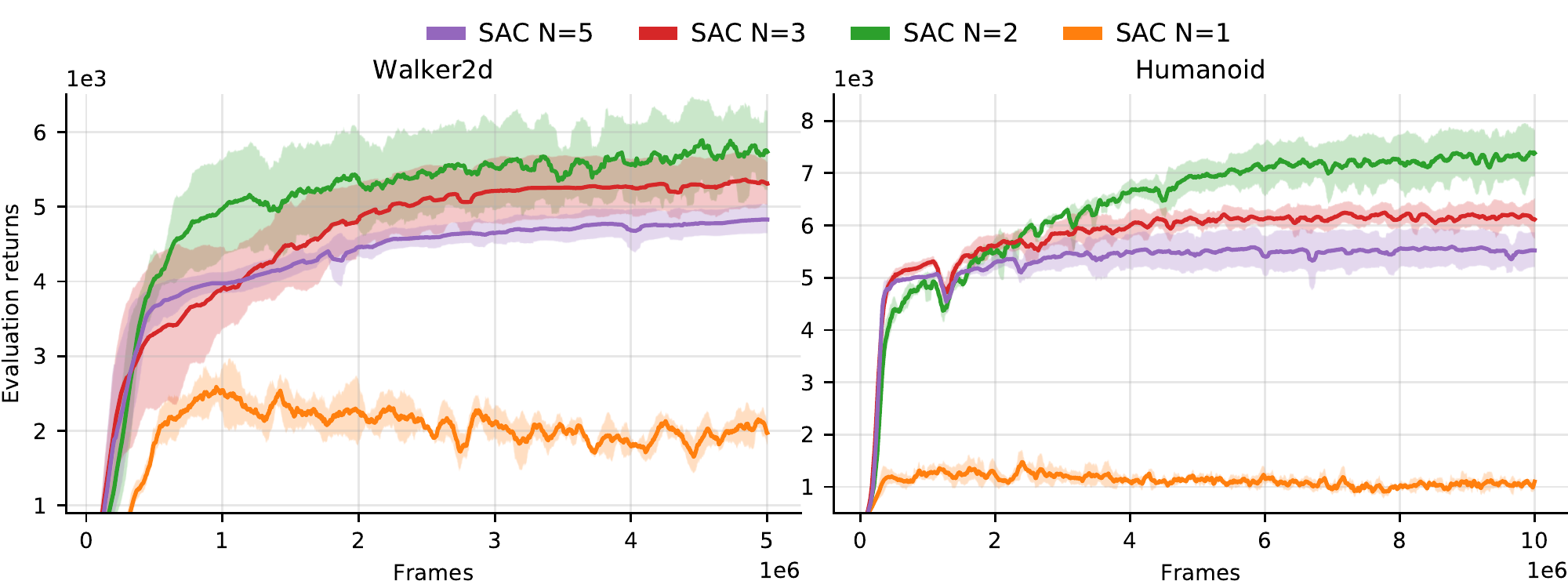}
\caption{
Varying the number of critic networks $N$ under the $\min$ operation of the Clipped Double Q-learning estimate for SAC. 
Each critic networks is $2$ layers deep with $256$ neurons in each layer (the same network structure as in SAC \cite{haarnoja2018applications}). 
Smoothed with a window of 100, $\pm$ std is plotted.
}
\label{fig:n_critics_sac}
\end{figure*}

\begin{figure*}[h]
\centering
\includegraphics[width=0.8\textwidth]{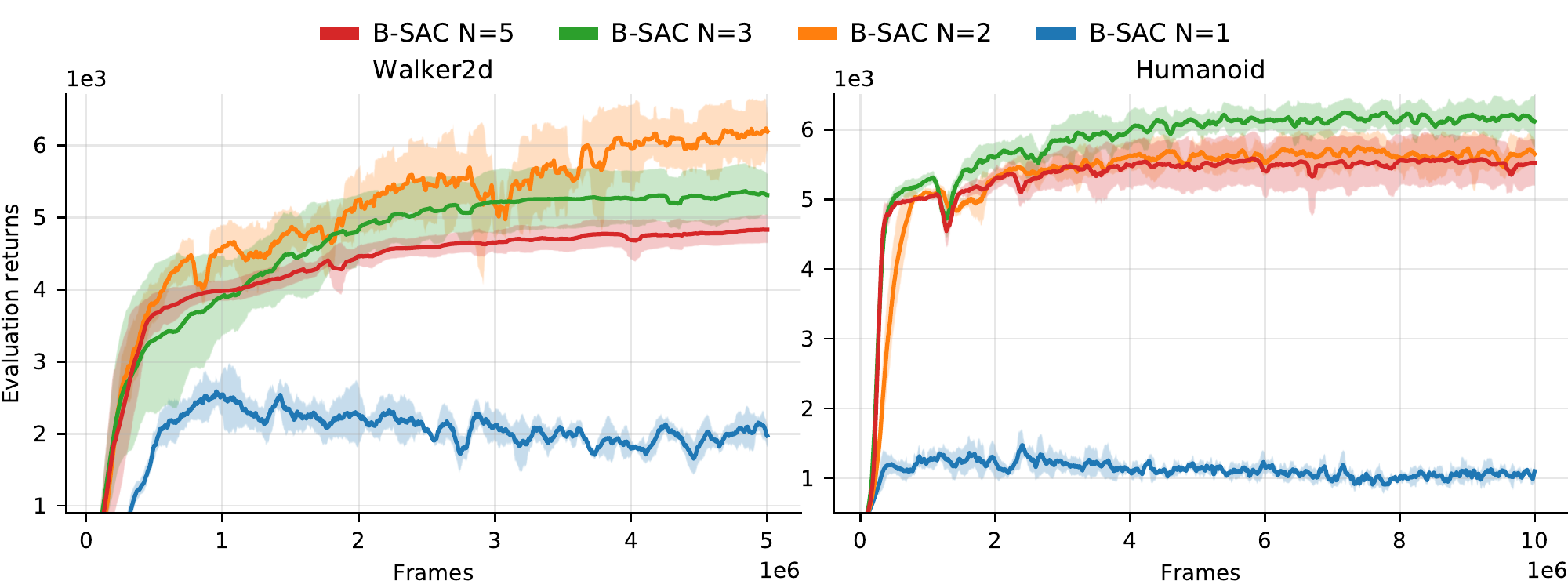}
\caption{
Varying the number of critic networks $N$ under the $\min$ operation of the Clipped Double Q-learning estimate for SAC. 
Each critic networks is $3$ layers deep with $512$ neurons in each layer.
Smoothed with a window of 100, $\pm$ std is plotted.
}
\label{fig:n_critics_big_sac}
\end{figure*}

We have found that for MuJoCo benchmarks it is not possible to improve performance upon the published results by controlling the overestimation in such a coarse way for both the regular network size  (Figure \ref{fig:n_critics_sac}), and for the increased network size (Figure \ref{fig:n_critics_big_sac}).

%%%%%%%%%%%%%%%%%%%%%%%%%%%%%%%%%%%%%%%%%%%%%%%%%%%%%%%%%%%%%%%%%%%%%%%%%%%%%%%
%%%%%%%%%%%%%%%%%%%%%%%%%%%%%%%%%%%%%%%%%%%%%%%%%%%%%%%%%%%%%%%%%%%%%%%%%%%%%%%

\end{document}